\documentclass[sigconf]{acmart}

\renewcommand\footnotetextcopyrightpermission[1]{} % removes footnote with conference information in first column
\AtBeginDocument{%
  \providecommand\BibTeX{{%
    \normalfont B\kern-0.5em{\scshape i\kern-0.25em b}\kern-0.8em\TeX}}}

\setcopyright{acmcopyright}
\copyrightyear{2021}
\acmYear{2021}
\acmDOI{10.1145/1122445.1122456}

\acmConference[]{}
\acmBooktitle{}
\acmPrice{15.00}
\acmISBN{978-1-4503-XXXX-X/18/06}

\usepackage{hyperref}
\usepackage{url}

\usepackage{soul}
\usepackage[tableposition=top]{caption}

\usepackage{subcaption}
\usepackage{graphicx}

\newbool{showComments}
\booltrue{showComments}
\ifbool{showComments}{%
\newcommand{\mj}[1]{\sethlcolor{cyan}\hl{[Partha: #1]}}
\newcommand{\lo}[1]{\sethlcolor{yellow}\hl{[Lorena: #1]}}
\newcommand{\sh}[1]{\sethlcolor{lime}\hl{[Sangwon: #1]}}
\newcommand{\rdj}[1]{\sethlcolor{orange}\hl{[Rene: #1]}}
\newcommand{\del}[1]{\textcolor{red}{#1}}
}{
\newcommand{\mj}[1]{}
\newcommand{\lo}[1]{}
\newcommand{\rdj}[1]{}
\newcommand{\sh}[1]{}
\newcommand{\del}[1]{}
}

\newcommand{\us}{\textit{Stochastic-Shield}}

\begin{document}

\title{Stochastic-Shield: A Probabilistic Approach Towards Training-Free Adversarial Defense in Quantized CNNs}

% \author{Anonymous authors}
% \affiliation{%
%   \institution{Paper under double-blind review}
% }
\author{Lorena Qendro}
\affiliation{%
  \institution{University of Cambridge}
}
\authornote{This work was done when the author was affiliated with Arm ML Research Lab.}

\author{Sangwon Ha}
\affiliation{%
  \institution{Arm ML Research Lab}
}

\author{René de Jong}
\affiliation{%
  \institution{Arm ML Research Lab}
}
\author{Partha Maji}
\affiliation{%
  \institution{Arm ML Research Lab}
}
\renewcommand{\shortauthors}{Qendro, et al.}

\begin{abstract}
Quantized neural networks (NN) are the common standard to efficiently deploy deep learning models on tiny hardware platforms. However, we notice that quantized NNs are as vulnerable to adversarial attacks as the full-precision models.
% However, we notice that they are vulnerable to adversarial attacks to a similar extent as 32-bit floating-point models.
% Deep learning models have great representation capability but they lack robustness in presence of adversarial attacks. 
With the proliferation of neural networks on small devices that we carry or surround us, there is a need for efficient models without sacrificing trust in the prediction in presence of malign perturbations. Current mitigation approaches often need adversarial training or are bypassed when the strength of adversarial examples is increased.

% Current mitigation approaches often need retraining or are bypassed when adversarial examples increase strength. 
In this work, we investigate how a probabilistic framework would assist in overcoming the aforementioned limitations for quantized deep learning models. We explore \us: a flexible defense mechanism that leverages \textit{input filtering} and a probabilistic deep learning approach materialized via \textit{Monte Carlo Dropout}. We show that it is possible to jointly achieve efficiency and robustness by accurately enabling each module without the burden of re-retraining or ad hoc fine-tuning.
\end{abstract}

%   <concept_desc>Human-centered computing~Ubiquitous and mobile computing</concept_desc>
%   <concept_desc>Computing methodologies~Artificial intelligence; Supervised learning by classification.</concept_desc>

% \ccsdesc[500]{Human-centered computing~Ubiquitous and mobile computing}
% \ccsdesc[500]{Computing methodologies~Artificial intelligence; Supervised learning by classification.}


\maketitle
\pagestyle{plain}

\section{Introduction}
\begin{figure*}[t]
    \centering
    \begin{subfigure}[t]{0.35\textwidth}
        \centering
        \includegraphics[width=\textwidth]{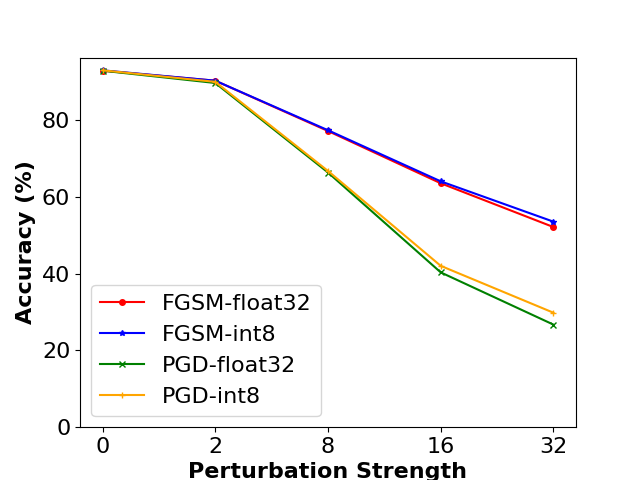}
        \caption{CIFAR10 on VGG16}
  \label{fig:vgg16_acc}
%   \vspace{-0.25in}
    \end{subfigure}
        \begin{subfigure}[t]{0.35\textwidth}
        \centering
        \includegraphics[width=\textwidth]{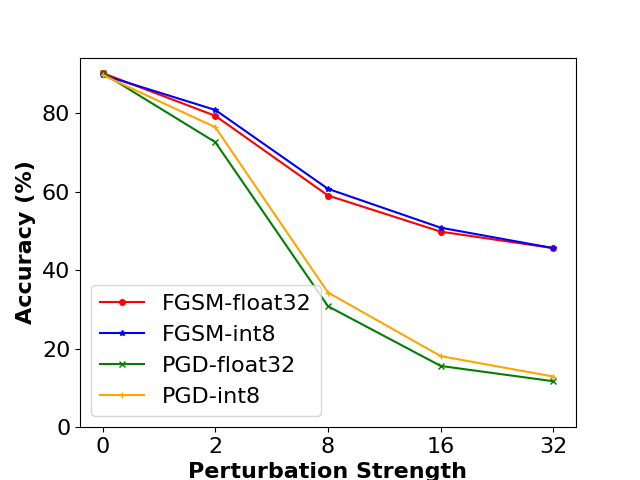}
        \caption{CIFAR10 on MobileNetV2}
  \label{fig:vgg16_acc}
%   \vspace{-0.25in}
    \end{subfigure}
    % \vspace{-0.10in}
    \caption{Accuracy under adversarial attack for floating-point precision and quantized neural networks. 8-bit integer-quantized NNs and 32-bit floating-point NNs are equally vulnerable to attacks. Setup: VGG16 and MobileNetV2 on CIFAR10 under FGSM and PGD (10 iterations) attacks.
    }
    \label{fig:fp-quant}
    % \vspace{-0.15in}
\end{figure*}

%  \vspace{-0.15in}
Quantized neural networks have facilitated the deployment of deep learning models on tiny devices in real-time applications by reducing computation and memory costs. However, these networks are as vulnerable to adversarial attacks as traditional non-optimized deep learning networks~\citep{lin2019defensive}. Although, the reduced bit-width should help denoising  small perturbations, there is negligible or no improvement to any perturbation strength from current adversarial attacks (see Figure \ref{fig:fp-quant}). This vulnerability is a  huge concern in safety-critical scenarios, such as health applications, autonomous driving, face and fingerprint identification, and voice recognition~\citep{amodei2016concrete}. 

The multiplicity of attacks grows continuously and traditional approaches that require retraining and re-deployment of the whole network are a significant burden and often not feasible. Adversarial perturbations do not stop at the feature level but they are propagated through the network, by easily bypassing techniques which focus on the feature space only~\citep{xu2017feature}. These attacks can be made virtually indistinguishable to human perception. However, they might be still visible to the human eye but not perceived as an attack (e.g. graffiti or stickers on road signs~\citep{li2019adversarial, lee2019physical}). Therefore, it is important to provide robustness and adversarial defense to real scenarios against attacks in a wide range of the strength scale.

To this purpose, we investigate how and to what extent can a probabilistic approach help this ecosystem. 
Through \us, we study how to obtain robustness against adversarial attacks operating at the feature and model parameter space via a modular adversarial defense which can be applied to already trained deep learning models. 
Moreover, since the majority of these models are and will be deployed on ubiquitous embedded devices which need a reduced computation demand by design, we consider 8-bit quantized deep learning models.

\us{} adopts an \textit{input filtering} layer to create a first shielding in addition to a probabilistic approach for more sophisticated and higher strength attacks. Input filtering consists of two blocks: input quantization and median smoothing which restrict the degrees of adversarial freedom in the input space. Additionally, the probabilistic framework via \textit{Monte Carlo Dropout}~\citep{gal2016dropout} introduces stochasticity to the network and, therefore, provides an implicit ensemble of models at runtime to make it more difficult for the attacker to aim at a specific set of neurons. Our experiments, which aim to measure robustness in presence of attacks at different strength levels, show that the use of the probabilistic framework becomes crucial when the attack strength increases and operating at both feature and models space we can achieve higher robustness for quantized models.
% \vspace{-0.20in}

\section{Background}
In this section, we navigate through common adversarial attacks, current mitigation approaches and, finally, we present the background on the probabilistic approach adopted in our framework.
\subsection{Adversarial Attacks}
Given a clean image $\mathbf{X}$, an adversarial attack introduces small perturbations $\nabla$ such that the class prediction for $\mathbf{X}$ and $\mathbf{X}^{adv}$ differs.

\textbf{Fast Gradient Sign Method (FGSM):} Introduced by~\citep{goodfellow2014explaining}, FGSM is a single step attack which aims to find the adversarial perturbations by moving in the opposite direction to the gradient of the loss function $L(\mathbf{X}, y)$ w.r.t. the image ($\nabla$):
% \vspace{-0.058in}
\begin{equation}
     \mathbf{X}^{adv} = \mathbf{X} + \epsilon * sign(\nabla_{\mathbf{X}}L(\mathbf{X}, y)),
\end{equation}
where $\epsilon$ is the step size which restricts the $l_{\infty}$ of the perturbation.

\textbf{Projected Gradient Descend (PGD):} A stronger variant of FGSM~\citep{kurakin2016adversarial} consists on applying it iteratively introducing a small step $\alpha$:
% \vspace{-0.058in}
\begin{equation}
    \mathbf{X}^{adv}_0 = \mathbf{X'}, \mathbf{X}^{adv}_{n+1} = clip^{\epsilon}_{\mathbf{X}} \{\mathbf{X}^{adv}_{n} + \alpha sign\big(\nabla_{\mathbf{X}}L\big(\mathbf{X}^{adv}_{n}, y \big) \big)\}
    \label{bim}
\end{equation}
where
% \vspace{-0.058in}
\begin{equation}
        \mathbf{X'} = \mathbf{X} + \epsilon_1 * sign\big(\mathcal{N}\big(\mathbf{0}^d,\mathbf{I}^d\big)\big)
    \label{rfsgm}
    % \vspace{-0.058in}
\end{equation}

is an additional prepended random step~\citep{tramer2017ensemble} which avoids going towards a false direction of ascent. Both steps, \ref{bim} and \ref{rfsgm}, make PGD~\citep{madry2017towards} which proves to be a universal first-order attack.

\textbf{Carlini Wagner (CW): } CW~\citep{carlini2017towards} is a very strong iterative attack which works with various $l_p$ norms. If we consider CW $L2$ which aims to minimize the following function:
% \vspace{-0.058in}
\begin{equation}
    \|\nabla\|^2_2 + c f\big(\mathbf{X} + \nabla\big)
    % \vspace{-0.058in}
\end{equation}

The constant $c$ is adjusted through line search in order to pick a point where $f\big(\mathbf{X} + \nabla\big)$ becomes negative and $f(\cdot)$ is defined as:
% \vspace{-0.058in}
\begin{equation}
    f(\mathbf{X}^{adv}) = Z(\mathbf{X}^{adv})_y - max\{Z(\mathbf{X}^{adv}_{y_a}) : y_a \ne y\}
    % \vspace{-0.058in}
\end{equation}

Where $Z(\mathbf{X}^{adv}_{y_a})$ gives the pre softmax predictions for class $y_a$ on the adversarial image $\mathbf{X}^{adv}$, instead, $y$ is the correct class which the attack wants to diverge from.

\begin{figure*}[ht]
\begin{center}
\includegraphics[width=1.0\textwidth]{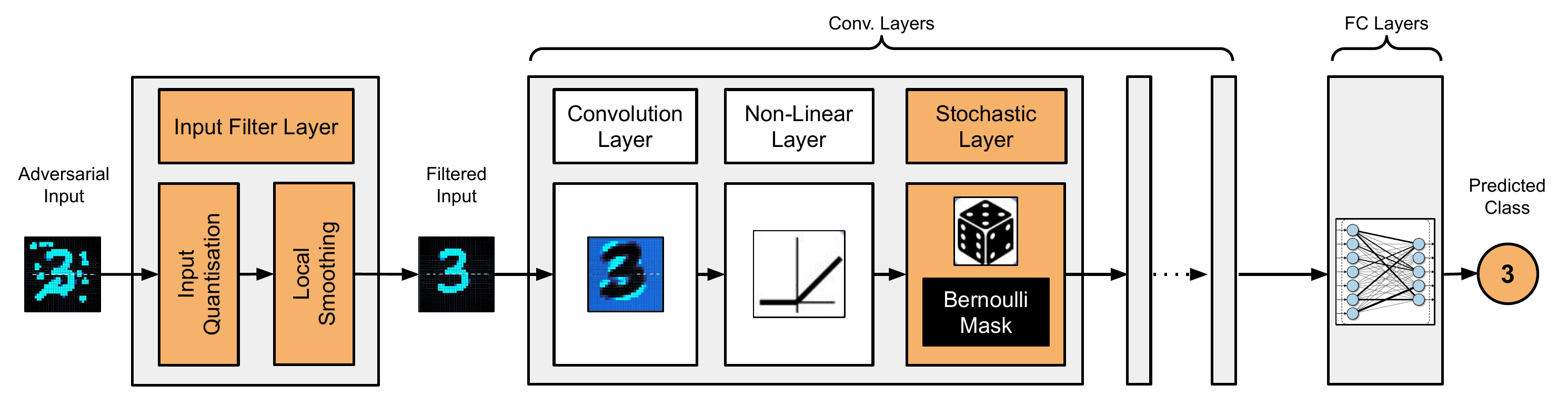}
    \caption{\us{} = Input filtering layer + A stochastic neural network (Bernoulli masks to the weight matrix via the dropout-based stochastic layer). Input filtering is performed only once, instead the model is run $T$ times to perform the forward stochastic passes needed for MCDrop.}
    \label{fig:ushield}
\end{center}
% \vspace{-0.12in}
\end{figure*}

\subsection{Adversarial detection and defense}
%  \vspace{-0.10in}
\textbf{Adversarial training}~\citep{goodfellow2014explaining, madry2017towards, zhang2019self} minimizes the risk that a perturbed sample can be misclassified, although its side-effects are not negligible. Classifiers trained with adversarial examples learn fundamentally different representations compared to standard classifiers reducing accuracy~\citep{tsipras2018robustness} or they can cause
disparity on accuracy for both clean  and adversarial samples between different classes~\citep{xu2020robust}. In addition, they are extremely resource-consuming
% , need to be trained on specific adversarial examples to guarantee robustness to them during inference, 
and cannot be applied to already deployed or trained network suggesting a very limited generalization and applicability in the real world.

\textbf{Feature Squeezing}~\citep{xu2017feature} aims to reduce the degrees of freedom available to an adversary by “squeezing” out unnecessary input features e.g. color-depth bit reduction. The technique consists on running the network with both the original and transformed input and detect the adversarial if the difference in the prediction passes a certain $\alpha$ threshold. Although, its computational overhead is minimal, it can be easily bypassed by increasing adversarial strength~\citep{sharma2018bypassing}). In \us{}, we adopt the promising transformations (input quantization and local smoothing) used in this approach. We use it differently as an additional filtering layer to the NNs for mitigation in reducing the chance of perturbed pixels to propagate in the quantized model.

\textbf{The taboo trap}~\citep{shumailov2018taboo} trains the model applying restrictions on activations, and during inference any sample violating these restrictions is considered adversarial. This solution is not generic as it would fail in presence of covariate shifts, and in addition, the attack can discover the limit of activations used in the restriction too.

\textbf{Ptolemy}~\citep{gan2020ptolemy} detects adversarial samples at runtime based on the observation that malign samples tend to activate distinctive paths from those of benign inputs. The main drawback of Ptolemy is having a backdoor in the model since the attack can learn which specific paths get activated by profiling too. In contrast, \us{} does not depend on any specific path, therefore,  profiling its activation path will give no information due to its stochastic properties.

\textbf{Defensive quantization (DQ)}~\citep{lin2019defensive} observes that the quantization operation amplifies the perturbation noise when passing through the deep NN. To overcome the aforementioned issue, this method proposes controlling the Lipschitz constant of the network during quantization, such that the magnitude of the adversarial noise remains non-expansive during inference. Nevertheless, DQ needs significant quantization-aware training and the quantization is applied to activations only, while the weights are still represented in 32-bit floating-point which is not a real world scenario.

\textbf{Stochastic approaches} assess adversarial attacks by looking at the likelihood of perturbed examples and the distance between perturbed and clean examples for each classifier~\citep{alemi2016deep, papernot2016distillation, li2017dropout}.~\citep{bradshaw2017adversarial} propose a Gaussian Process (GP) hybrid deep NN to help mitigating adversarial attacks, however, a closed-form solution for GPs has a $\mathcal{O}(n^3)$ computational complexity. In addition, GPs are very sensitive to quantization and thus implementation on embedded hardware is challenging. A promising direction, Monte Carlo Dropout (MCDrop)~\citep{gal2015bayesian}, casts dropout training as approximate inference in Bayesian CNNs.~\citep{feinman2017detecting} investigate model confidence on adversarial samples through MCDrop by looking at uncertainty estimates. They operate a two-feature approach similar to ours, however, they need to observe the density estimates during training to calculate to detect the points that lie far from the data manifold, which might lead to issues adapting to data shifts. While these techniques study the uncertainty for adversarial detection, we envision using the probabilistic framework to prevent the attack from changing the prediction by making it more difficult for the attacker to overcome the randomness introduced by dropout and finding a deterministic path towards the wrong class.

\subsection{Monte Carlo Dropout}
Standard dropout~\citep{srivastava2014dropout} was initially introduced as a regularization technique to avoid overfitting. It consists of dropping random units with a certain probability $p$ which allows a shift from a deterministic model to a stochastic one constituted of implicit ensemble of networks. Therefore, a fully connected layer with dropout is formulated as:
{
\begin{align}
\begin{split}
 \textbf{z}_{[i]}^{(l)}  &\sim  \mathrm{Bernoulli}\big(\cdot|\textbf{p}_{[i]}^{(l)}\big)
\\
 \tilde{\textbf{W}}^{(l)}  &=  \mathrm{diag}\big(\textbf{z}^{(l)}\big)\textbf{W}^{(l)}
\\
 \textbf{y}^{(l)}  &=  \textbf{x}^{(l)}\tilde{\textbf{W}}^{(l)} + \textbf{b}^{(l)}
\\
 \textbf{x}^{(l+1)}  &=  f^{(l)}\big(\textbf{y}^{(l)}\big)
\end{split}
\end{align}
}
where $\textbf{x}^{(l)}$ and $\textbf{y}^{(l)}$ are the input and output of that layer, and  $\textbf{f}^{(l)}(\cdot)$ is the nonlinear activation function. $\textbf{W}^{(l)}$ is the weight matrix of \textit{l} with dimensions $\textit{K}^{(l)}$ x $\textit{K}^{(l-1)}$ and $\textbf{b}^{(l)}$ is the bias vector of dimensions $\textit{K}^{(l)}$. Here $\textbf{z}_{[i]}^{(l)}$ are Bernoulli distributed random variables with some probabilities $\textbf{p}_{[i]}^{(l)}$. The $\mathrm{diag}(\cdot)$ maps vectors to diagonal matrices.

~\citep{gal2016dropout} proved the equivalence between dropout training in a neural network and approximate inference in a deep Gaussian Process. They showed that the objective converges to a minimization of the Kullback-Leibler divergence between an approximate distribution and the posterior of a deep Gaussian process marginalized over its covariance function parameters. The true posterior distribution is, therefore, approximated by the variational distribution $q\big(\tilde{\textbf{W}}^{(l)}\big)$ where $\tilde{\textbf{W}}^{(l)}$ represents the random variables used in dropout operations as described in (1). 
{
\begin{align}
\begin{split}
 \textbf{z}_{[i]}^{(l)} &\sim  \mathrm{Bernoulli}\big(\cdot|\textbf{p}_{[i]}^{(l)}\big)
\\
 q\big(\tilde{\textbf{W}}^{(l)}\big) &=  \mathrm{diag}\big(\textbf{z}^{(l)}\big)\textbf{W}^{(l)}
\end{split}
\end{align}
}

Finally, they perform Monte Carlo (MC) sampling of the random variables $\mathcal{W}$,

% \vspace{-0.10in}
{
\begin{align}
     q(\textbf{y}|\textbf{x}) =  \frac{1}{T} \sum_{t=1}^{T}q(\textbf{y}|\textbf{x}, \mathcal{W}_t),
\end{align}
}

where T is the number of MC samples. This method is called Monte Carlo Dropout (MCDrop) and is equivalent to performing  T stochastic passes.
Similarly for convolution layers~\citep{gal2015bayesian}, the sampled Bernoulli random variables $\textbf{z}_{i,j,k}$ are applied as masks to the weight matrix $\textbf{W}_i \cdot diag([\textbf{z}_i,j,k])$ which is equivalent to setting weights to $0$ for different elements of the input.
\section{Stochastic-Shield}
% \vspace{-0.12in}

% Clues and limitations from previous work inspired our exploration on a probabilistic approach towards adversarial defense for quantized neural networks. 
Observed data can be consistent with many models, and therefore, which model is appropriate given the data, is uncertain. Adversarial attacks increase the uncertainty by adding perturbations that trick the model to change the prediction class.  A probabilistic framework can capture the uncertainty in the data and model parameters, which are spaces where the adversarial attacks particularly aim to operate. Therefore, our intuition led to exploring \us{}, to offer robustness against adversarial attacks operating at the input and model parameter space.

\begin{figure*}[t]
    \centering
    \begin{subfigure}[t]{0.32\textwidth}
        \centering
        \includegraphics[width=\textwidth]{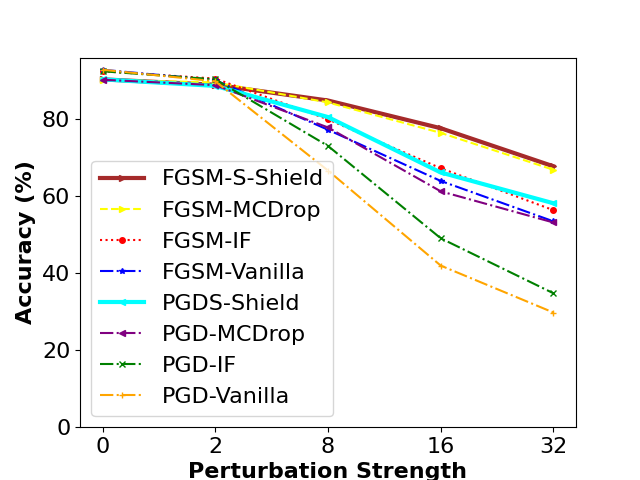}
        \caption{CIFAR10 on VGG16}
  \label{fig:vgg16_fs_mc_acc}
%   \vspace{-0.25in}
    \end{subfigure}
        \begin{subfigure}[t]{0.32\textwidth}
        \centering
        \includegraphics[width=\textwidth]{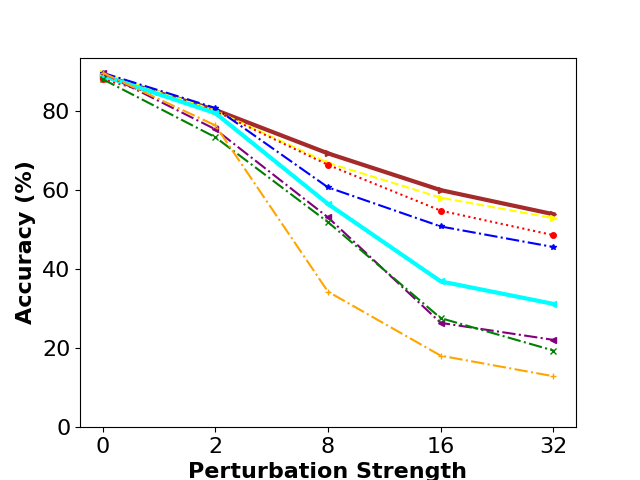}
        \caption{CIFAR10 on MobileNetV2}
                \label{fig:mv2_fs_mc_acc}

    \end{subfigure}
        \begin{subfigure}[t]{0.32\textwidth}
        \centering
        \includegraphics[width=\textwidth]{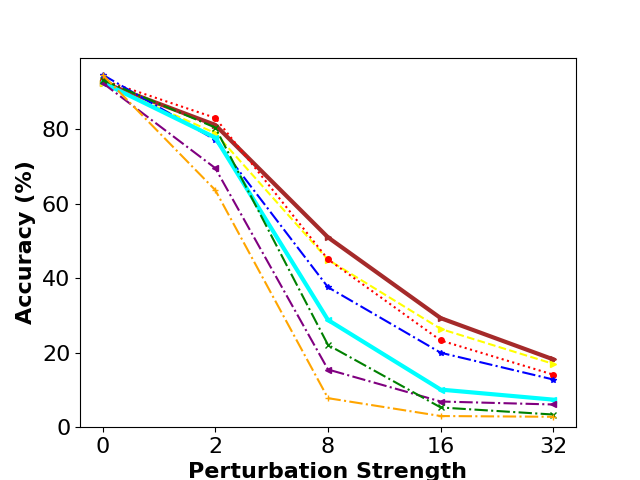}
        \caption{SVHN on CNN}
  \label{fig:cnn_fs_mc_acc}
%   \vspace{-0.25in}
    \end{subfigure}
    
    % \vspace{-0.10in}
    \caption{Accuracy on clean ($\epsilon=0$) and perturbed input for FGSM and PGD attacks with different perturbation strength for Vanilla (no defence) DNN, IF (Input Filter, no\_bits=5, median\_filter=2x2), MCDrop5 (5 samples), and \us{}. All models are quantized with 8-bit weights and activations. 
    }
    \label{fig:accuracy}
    % \vspace{-0.15in}
\end{figure*}
\us{} (Figure  \ref{fig:ushield}) initially adopts an \textit{input filtering} layer to create a first shielding which aims to filter out some of the perturbations before they are propagated in the network.
% and a probabilistic approach for more sophisticated and higher strength adversarial attacks. 
Input filtering, which represents quantization and median smoothing in the input space, can capture the perturbations and offers a solution to reduce the degrees of freedom available to the adversary by filtering out unnecessary features.
This layer can reduce the effect of the perturbations in the input but it cannot help against stronger attacks which are propagated in the deeper layers. 

To solve the latter, we adopt Monte Carlo Dropout (MCDrop), which allows to create an implicit ensemble of quantized models by running the same model multiple times with dropout layers activated during inference. Compared to deterministic deep ensembles which, once trained, are fixed; the stochasticity via MCDrop helps mitigating the adversarial attack by making it more difficult to profile the activation path and attack the implicit ensemble as a whole.
While operating in two different core parts of the network, we observe these two techniques together to be very effective on detecting and mitigating adversarial attacks for 8-bit integer-quantized deep learning models.

%  \us{}  offers robustness against adversarial attacks operating at the feature and model parameter space. It adopts \textit{feature squeezing} to create a first shielding and a probabilistic approach for more sophisticated and higher strength adversarial attacks. Our approach is flexible and adaptable to the energy envelope as both shielding levels can be applied separately or cooperatively to provide the best defense. \us{} uses Monte Carlo Dropout to introduce stochasticity to the network and, therefore, provide an implicit ensemble of models at runtime to make it more difficult for the attacker to follow one specific path. We evaluate our approach with different state-of-the-art architectures and datasets to show that it can be generalizable to different tasks and architectures and robust against various adversarial attacks at different strength levels. Our experiments show that the use of the probabilistic framework becomes crucial when the attack strength increases and operating at both feature and models space we can achieve the desired robustness. \lo{needs to be boosted more here}

To answer our initial question on how and to what extent can a probabilistic approach help mitigating adversarial attacks in quantized models, we test \us{} on 3 networks: MobileNetV2 and  VGG16 for CIFAR10 and a CNN based one for Street View House Number (SVHN) (See \ref{implementation} for details). For each of the networks and datasets, we measure the accuracy and the expected calibration error (ECE).

\begin{figure*}[t]
    \centering
    \begin{subfigure}[t]{0.32\textwidth}
        \centering
        \includegraphics[width=\textwidth]{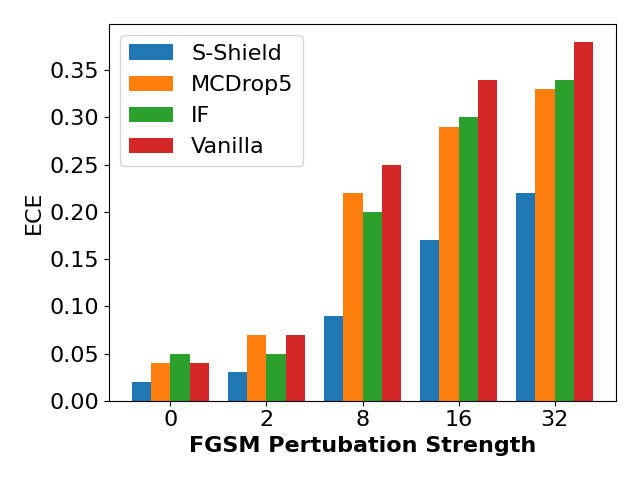}
        \label{fig:mv2_ece_fsgm}
        % \caption{MobileNetV2}
          \vspace{-0.25in}
    \end{subfigure}
    % \hspace{0.02\textwidth}
    \begin{subfigure}[t]{0.32\textwidth}
        \centering
        \includegraphics[trim=0 0 0 0,clip,width=\textwidth]{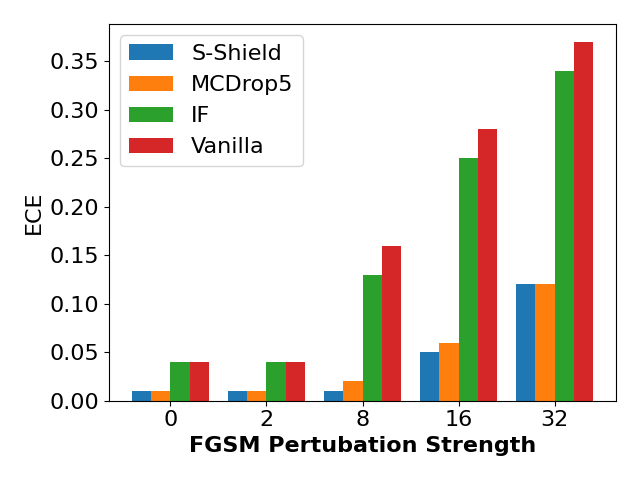}
        \label{fig:vgg16_ece_fsgm}
        % \caption{{FGSM on VGG16}}
          \vspace{-0.25in}
    \end{subfigure}
            \begin{subfigure}[t]{0.32\textwidth}
        \centering
        \includegraphics[trim=0 0 0 0,clip,width=\textwidth]{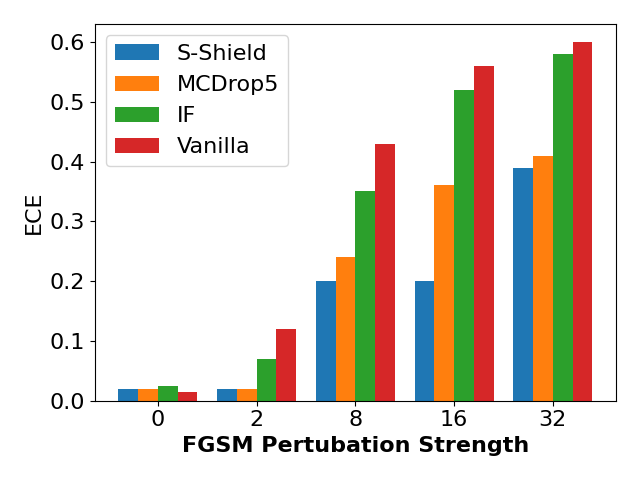}
        \label{fig:cnn_ece_fsgm}
        % \caption{{FGSM on CNN}}
        \vspace{-0.25in}
    \end{subfigure}
            \begin{subfigure}[t]{0.32\textwidth}
        \centering
        \includegraphics[trim=0 0 0 0,clip,width=\textwidth]{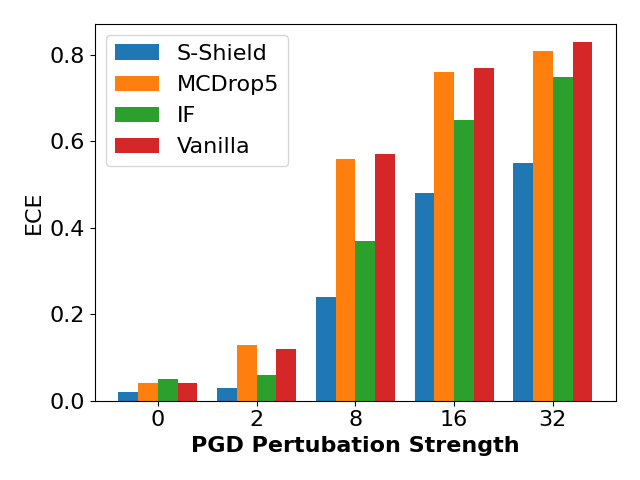}
        \label{fig:mv2_ece_pgd}
        \vspace{-0.25in}
         \caption{{MobileNetV2}}
         
    \end{subfigure}
    \begin{subfigure}[t]{0.32\textwidth}
        \centering
        \includegraphics[trim=0 0 0 0,clip,width=\textwidth]{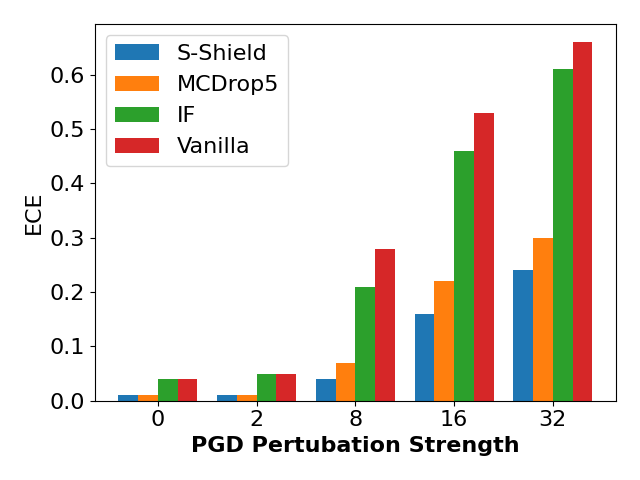}
        \label{fig:vgg16_ece_pgd}
                \vspace{-0.25in}
        \caption{{VGG16}}
    \end{subfigure}
    \begin{subfigure}[t]{0.32\textwidth}
        \centering
        \includegraphics[trim=0 0 0 0,clip,width=\textwidth]{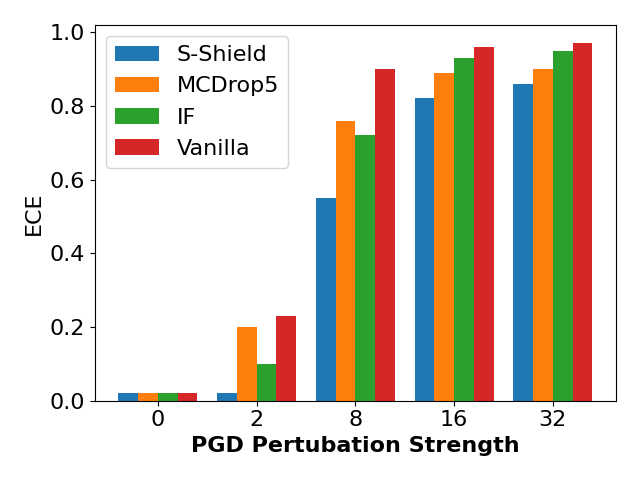}
        \label{fig:cnn_ece_pgd}
        \vspace{-0.25in}
        \caption{{CNN}}
    \end{subfigure}
    % \vspace{-0.10in}
    \caption{Expected calibration error for Vanilla (no defence) DNN, IF (Input Filter, no\_bits=5, median\_filter=2x2), MC5 (MCDrop 5 samples), and \us{}.
    % Lower error indicates better calibrated model, less prone to falling into the adversarial trap.
    }
    \label{fig:all}
    % \vspace{-0.15in}
\end{figure*}

Figure \ref{fig:accuracy} indicates the accuracy of the pre-trained 8-bit quantized models under various adversarial attacks (FGSM and PGD) and strengths. The mitigation techniques are applied separately and together to show the efficacy of each individually and in unison. We see that input filtering is able to keep an acceptable accuracy when the adversarial strength is low, but when the attacks are stronger it continuously fails to achieve the results of MCDrop with 5 samples. However, in all cases combination of both techniques that make \us{} achieve the desired robustness presenting a considerably higher accuracy (improvements up to 3--folds compared to Vanilla models and up to 2--folds compared to input filtering only) in presence of attacks and better calibrated  models (see Figure~\ref{fig:all}). Higher the adversarial strength, bigger the need for the probabilistic approach and \us{}.

Table~\ref{tab:cw_results} shows the results when using a CW L2 non-targeted attack reinforcing this finding. Although our approach is the best performing, the difference to the other approaches is not big since the attack is not very strong (compared to PGD($\epsilon=32$) e.g. which drops the accuracy to less than 10\%). CW is very costly for embedded devices, however, a CW $L_\infty$ attack would have a greater effect on accuracy and \us{} would be a must to mitigate it.

% \vspace{-0.10in}
% \vspace{-0.10in}

\begin{table}[!h]
\huge
    \centering
    \caption{Accuracy and Expected Calibration Error (ECE) of models under CW L2 attack with default parameters from~\citep{carlini2017towards}.}
    \resizebox{0.49\textwidth}{!}{
    \begin{tabular}{c c c c c c c c c}\\
    & \multicolumn{2}{c}{Vanilla} & \multicolumn{2}{c}{Input Filter} & \multicolumn{2}{c}{MCDrop 5} & \multicolumn{2}{c}{Stochastic-Shield}\\
     & \textit{Acc} & \textit{ECE} & \textit{Acc} & \textit{ECE} & \textit{Acc} & \textit{ECE} & \textit{Acc} & \textit{ECE}\\
     \hline\hline
    MobileNetV2& 51.9\% & 0.75 & 52.5\% & 0.74 & 53.3\% & \textbf{0.73} & \textbf{54.1\%} & 0.75\\
    VGG16& 66.7\% & 0.8 & 65.7\% & 0.8 & 70.1\% & 0.73 & \textbf{70.7\%} & \textbf{0.73}\\
    CNN& 69.7\% & \textbf{0.68} & 80.8\% & 0.75 & 83.6\% & 0.71 & \textbf{83.6\%} & 0.7\\
    \hline
    \end{tabular}}
    \label{tab:cw_results}
    % \vspace{-0.15in}
\end{table}
\section{Implementation Details}
\label{implementation}
We consider 3 networks: MobileNetV2 and  VGG16 for CIFAR10 and a CNN based one for Street View House Number (SVHN). All networks have been trained with dropout layers which is kept activated at inference time keeping its training dropout rate. The training does not include any adversarial training or hyper-parameter tuning to increase robustness.
The already trained networks are quantized to use 8-bit weights and activations via TensorFlow Model Optimization Toolkit~\citep{TMO}.

We implement the adversarial attacks (FGSM, PGD, and CW) in Keras using the library \textit{adversarial-robustness-toolbox}~\citep{nicolae2018adversarial}. For FGSM and PGD, we set the perturbation strength $\epsilon = \{0,2,8,16,32 \}$ and colors are represented from 0-255. PGD is an iterative attack and in our experiments we use 10 iterations. We use the L2 norm CW attack which aims to minimize the objective function by using the gradient descent. We used 1000 iterations, 9 binary search steps and confidence 0.0 as in~\citep{carlini2017towards}. These adversarial perturbations are applied to the whole test set and all the metrics represent the overall accuracy when an adversarial attack is applied to each sample.

For all experiments, we are performing 5 forward passes for MCDrop (MC5). This choice was made considering previous work~\citep{ovadia2019can, azevedo2020stochastic, qendro2021benefit} and the fact that although less computation heavy than pure Bayesian approaches or Gaussian Processes, MCDrop can introduce some overhead to the system given the multiple forward passes. Moreover, keeping a low number of samples allows us to see what the technique can achieve keeping the overhead at a minimum.

\begin{table}[ht]
\Huge
    \centering
    \caption{Architecture and implementation details of the considered deep learning models.}
    \resizebox{0.49\textwidth}{!}{
    \begin{tabular}{c c  c c  c c }
    \hline
    \textbf{VGG16} & CIFAR10 & \textbf{MobileNetV2} & CIFAR10 & \textbf{CNN based} & SVHN\\
    \hline \hline
    Layer & Details & Layer & Details &Layer & Details\\ [0.5ex]
    \hline
    Conv2D(BN,ReLU)    &  3x3x64 & Conv2d(BN, ReLU) & 3x3x32 & Conv2d(BN, ReLU) & 3x3x32\\
    Dropout & 0.3 & InvertedResidual & 16 & Conv2d(BN, ReLU) & 3x3x32\\
    Conv2D(BN,ReLU)    &  3x3x64 & InvertedResidual (x2) & 24 & MaxPooling2D & 2x2\\
    MaxPooling2D & 2x2 & InvertedResidual (x3) & 32 & Dropout & 0.3\\
    Conv2D(BN,ReLU)    &  3x3x128  & InvertedResidual (x4) & 64 & Conv2d(BN, ReLU) & 3x3x64\\
    Dropout & 0.4 & Dropout & 0.25 & Conv2d(BN, ReLU) & 3x3x64\\
    Conv2D(BN,ReLU)    &  3x3x128 & InvertedResidual (x3) & 96 & MaxPooling2D & 2x2\\
    MaxPooling2D & 2x2 & Dropout & 0.25 & Dropout & 0.3\\
    
    Conv2D(BN,ReLU)    &  3x3x256 & InvertedResidual (x3) & 160 & Conv2d(BN, ReLU) & 3x3x128\\
    Dropout & 0.4 & Dropout & 0.25 & Conv2d(BN, ReLU) & 3x3x128\\
    Conv2D(BN,ReLU)    &  3x3x256 & InvertedResidual & 320 & MaxPooling2D & 2x2\\
    Dropout & 0.4 & Dropout & 0.25 & Dropout & 0.3\\
    Conv2D(BN,ReLU)    &  3x3x256 & Conv2D(BN, ReLU) & & Flatten &\\
    MaxPooling2D & 2x2 & GlobalAveragePooling2D & &Dense(ReLU) & 512\\
    Conv2D(BN,ReLU)    &  3x3x512 & Dense & 10 & Dropout & 0.3\\
    Dropout & 0.4 & Softmax & & Dense & 10\\
    Conv2D(BN,ReLU)    &  3x3x512 & && Softmax\\
    Dropout & 0.4 & \textit{where} &\\
    Conv2D(BN,ReLU)    &  3x3x512 &\textbf{InvertedResidual} & (filters)\\
    MaxPooling2D & 2x2 & \textit{is composed of} &\\
    Conv2D(BN,ReLU)    &  3x3x512 & Conv2D(BN, ReLU)&\\
    Dropout & 0.4 & DepthwiseConv2D(BN, ReLU)& \\
    Conv2D(BN,ReLU)    &  3x3x512 & Conv2D(BN)&\\
    Dropout & 0.4 &&\\
    Conv2D(BN,ReLU)    &  3x3x512 && \\
    MaxPooling2D & 2x2 &&\\
    Dropout & 0.5 &&\\
    Flatten & &&\\
    Dense & 512 &&\\
    Dropout & 0.5 &&\\
    Dense & 10 &&\\
    Softmax & &&\\
    \hline
    \end{tabular}}
    \label{tab:archs}
    % \vspace{-0.25in}
\end{table}
\section{Discussion and Future work}
% \vspace{-0.12in}

Quantized deep learning models are very susceptible to adversarial attacks questioning the trust in them especially when dealing with safety-critical applications. In this work, we investigate to what extent a probabilistic approach can help mitigating adversarial examples. To this purpose, we build \us{} a training-free multi-module shielding methodology which consists on adopting \textit{input filtering} and \textit{Monte Carlo Dropout} to provide models which are better calibrated and less sensitive to adversarial attacks. 

Our experiments show that although each module contributes in mitigating attacks at different perturbation strength, when used together they consistently deliver the best performance on 8-bit quantized neural networks. This framework is easily parallelizable by running the independent Monte Carlo samples simultaneously creating a batch of the same input during inference or by running each of the forward pass on different devices of a federated system. This parallelization would guaranty a minimal delay in latency and offer robustness even on the tiniest devices.

An interesting avenue for future work could focus on studying how an approach like \us{} would aid the robustness of models quantized with less than 8 bits (uniform 4-bit or mixed precision 8/4bit) models for even tinier devices that are used in IoT applications. Moreover, the findings in this work open a lot of opportunities in studying the effect of the level of randomness or dropout rate in adversarial mitigation.
% as well as the combination with explicit deep ensemble of networks~\citep{lakshminarayanan2016simple} to add another level of complexity increasing the cost of the attack.

\bibliographystyle{ACM-Reference-Format}
\bibliography{bibliography}


\begin{thebibliography}{29}

%%% the Web address.

\ifx \showCODEN    \undefined \def \showCODEN     #1{\unskip}     \fi
\ifx \showDOI      \undefined \def \showDOI       #1{#1}\fi
\ifx \showISBNx    \undefined \def \showISBNx     #1{\unskip}     \fi
\ifx \showISBNxiii \undefined \def \showISBNxiii  #1{\unskip}     \fi
\ifx \showISSN     \undefined \def \showISSN      #1{\unskip}     \fi
\ifx \showLCCN     \undefined \def \showLCCN      #1{\unskip}     \fi
\ifx \shownote     \undefined \def \shownote      #1{#1}          \fi
\ifx \showarticletitle \undefined \def \showarticletitle #1{#1}   \fi
\ifx \showURL      \undefined \def \showURL       {\relax}        \fi
% invisible to TeX
\providecommand\bibfield[2]{#2}
\providecommand\bibinfo[2]{#2}
\providecommand\natexlab[1]{#1}
\providecommand\showeprint[2][]{arXiv:#2}

\bibitem[\protect\citeauthoryear{??}{TMO}{[n.d.]}]%
        {TMO}
 \bibinfo{year}{[n.d.]}\natexlab{}.
\newblock \bibinfo{title}{TensorFlow Model Optimization}.
\newblock
  \bibinfo{howpublished}{\url{https://www.tensorflow.org/model_optimization/guide/quantization/training}}.
\newblock


\bibitem[\protect\citeauthoryear{Alemi, Fischer, Dillon, and Murphy}{Alemi
  et~al\mbox{.}}{2016}]%
        {alemi2016deep}
\bibfield{author}{\bibinfo{person}{Alexander~A Alemi}, \bibinfo{person}{Ian
  Fischer}, \bibinfo{person}{Joshua~V Dillon}, {and} \bibinfo{person}{Kevin
  Murphy}.} \bibinfo{year}{2016}\natexlab{}.
\newblock \showarticletitle{Deep variational information bottleneck}.
\newblock \bibinfo{journal}{\emph{arXiv preprint arXiv:1612.00410}}
  (\bibinfo{year}{2016}).
\newblock


\bibitem[\protect\citeauthoryear{Amodei, Olah, Steinhardt, Christiano,
  Schulman, and Man{\'e}}{Amodei et~al\mbox{.}}{2016}]%
        {amodei2016concrete}
\bibfield{author}{\bibinfo{person}{Dario Amodei}, \bibinfo{person}{Chris Olah},
  \bibinfo{person}{Jacob Steinhardt}, \bibinfo{person}{Paul Christiano},
  \bibinfo{person}{John Schulman}, {and} \bibinfo{person}{Dan Man{\'e}}.}
  \bibinfo{year}{2016}\natexlab{}.
\newblock \showarticletitle{Concrete problems in AI safety}.
\newblock \bibinfo{journal}{\emph{arXiv preprint arXiv:1606.06565}}
  (\bibinfo{year}{2016}).
\newblock


\bibitem[\protect\citeauthoryear{Azevedo, de~Jong, and Maji}{Azevedo
  et~al\mbox{.}}{2020}]%
        {azevedo2020stochastic}
\bibfield{author}{\bibinfo{person}{Tiago Azevedo}, \bibinfo{person}{Ren{\'e} de
  Jong}, {and} \bibinfo{person}{Partha Maji}.} \bibinfo{year}{2020}\natexlab{}.
\newblock \showarticletitle{Stochastic-YOLO: Efficient Probabilistic Object
  Detection under Dataset Shifts}.
\newblock \bibinfo{journal}{\emph{arXiv preprint arXiv:2009.02967}}
  (\bibinfo{year}{2020}).
\newblock


\bibitem[\protect\citeauthoryear{Bradshaw, Matthews, and Ghahramani}{Bradshaw
  et~al\mbox{.}}{2017}]%
        {bradshaw2017adversarial}
\bibfield{author}{\bibinfo{person}{John Bradshaw}, \bibinfo{person}{Alexander G
  de~G Matthews}, {and} \bibinfo{person}{Zoubin Ghahramani}.}
  \bibinfo{year}{2017}\natexlab{}.
\newblock \showarticletitle{Adversarial examples, uncertainty, and transfer
  testing robustness in Gaussian process hybrid deep networks}.
\newblock \bibinfo{journal}{\emph{arXiv preprint arXiv:1707.02476}}
  (\bibinfo{year}{2017}).
\newblock


\bibitem[\protect\citeauthoryear{Carlini and Wagner}{Carlini and
  Wagner}{2017}]%
        {carlini2017towards}
\bibfield{author}{\bibinfo{person}{Nicholas Carlini} {and}
  \bibinfo{person}{David Wagner}.} \bibinfo{year}{2017}\natexlab{}.
\newblock \showarticletitle{Towards evaluating the robustness of neural
  networks}. In \bibinfo{booktitle}{\emph{2017 ieee symposium on security and
  privacy (sp)}}. IEEE, \bibinfo{pages}{39--57}.
\newblock


\bibitem[\protect\citeauthoryear{Feinman, Curtin, Shintre, and Gardner}{Feinman
  et~al\mbox{.}}{2017}]%
        {feinman2017detecting}
\bibfield{author}{\bibinfo{person}{Reuben Feinman}, \bibinfo{person}{Ryan~R
  Curtin}, \bibinfo{person}{Saurabh Shintre}, {and} \bibinfo{person}{Andrew~B
  Gardner}.} \bibinfo{year}{2017}\natexlab{}.
\newblock \showarticletitle{Detecting adversarial samples from artifacts}.
\newblock \bibinfo{journal}{\emph{arXiv preprint arXiv:1703.00410}}
  (\bibinfo{year}{2017}).
\newblock


\bibitem[\protect\citeauthoryear{Gal and Ghahramani}{Gal and
  Ghahramani}{2015}]%
        {gal2015bayesian}
\bibfield{author}{\bibinfo{person}{Yarin Gal} {and} \bibinfo{person}{Zoubin
  Ghahramani}.} \bibinfo{year}{2015}\natexlab{}.
\newblock \showarticletitle{Bayesian convolutional neural networks with
  Bernoulli approximate variational inference}.
\newblock \bibinfo{journal}{\emph{arXiv preprint arXiv:1506.02158}}
  (\bibinfo{year}{2015}).
\newblock


\bibitem[\protect\citeauthoryear{Gal and Ghahramani}{Gal and
  Ghahramani}{2016}]%
        {gal2016dropout}
\bibfield{author}{\bibinfo{person}{Yarin Gal} {and} \bibinfo{person}{Zoubin
  Ghahramani}.} \bibinfo{year}{2016}\natexlab{}.
\newblock \showarticletitle{Dropout as a bayesian approximation: Representing
  model uncertainty in deep learning}. In
  \bibinfo{booktitle}{\emph{international conference on machine learning}}.
  PMLR, \bibinfo{pages}{1050--1059}.
\newblock


\bibitem[\protect\citeauthoryear{Gan, Qiu, Leng, Guo, and Zhu}{Gan
  et~al\mbox{.}}{2020}]%
        {gan2020ptolemy}
\bibfield{author}{\bibinfo{person}{Yiming Gan}, \bibinfo{person}{Yuxian Qiu},
  \bibinfo{person}{Jingwen Leng}, \bibinfo{person}{Minyi Guo}, {and}
  \bibinfo{person}{Yuhao Zhu}.} \bibinfo{year}{2020}\natexlab{}.
\newblock \showarticletitle{Ptolemy: Architecture support for robust deep
  learning}. In \bibinfo{booktitle}{\emph{2020 53rd Annual IEEE/ACM
  International Symposium on Microarchitecture (MICRO)}}. IEEE,
  \bibinfo{pages}{241--255}.
\newblock


\bibitem[\protect\citeauthoryear{Goodfellow, Shlens, and Szegedy}{Goodfellow
  et~al\mbox{.}}{2014}]%
        {goodfellow2014explaining}
\bibfield{author}{\bibinfo{person}{Ian~J Goodfellow}, \bibinfo{person}{Jonathon
  Shlens}, {and} \bibinfo{person}{Christian Szegedy}.}
  \bibinfo{year}{2014}\natexlab{}.
\newblock \showarticletitle{Explaining and harnessing adversarial examples}.
\newblock \bibinfo{journal}{\emph{arXiv preprint arXiv:1412.6572}}
  (\bibinfo{year}{2014}).
\newblock


\bibitem[\protect\citeauthoryear{Kurakin, Goodfellow, and Bengio}{Kurakin
  et~al\mbox{.}}{2016}]%
        {kurakin2016adversarial}
\bibfield{author}{\bibinfo{person}{Alexey Kurakin}, \bibinfo{person}{Ian
  Goodfellow}, {and} \bibinfo{person}{Samy Bengio}.}
  \bibinfo{year}{2016}\natexlab{}.
\newblock \showarticletitle{Adversarial machine learning at scale}.
\newblock \bibinfo{journal}{\emph{arXiv preprint arXiv:1611.01236}}
  (\bibinfo{year}{2016}).
\newblock


\bibitem[\protect\citeauthoryear{Lee and Kolter}{Lee and Kolter}{2019}]%
        {lee2019physical}
\bibfield{author}{\bibinfo{person}{Mark Lee} {and} \bibinfo{person}{Zico
  Kolter}.} \bibinfo{year}{2019}\natexlab{}.
\newblock \showarticletitle{On physical adversarial patches for object
  detection}.
\newblock \bibinfo{journal}{\emph{arXiv preprint arXiv:1906.11897}}
  (\bibinfo{year}{2019}).
\newblock


\bibitem[\protect\citeauthoryear{Li, Schmidt, and Kolter}{Li
  et~al\mbox{.}}{2019}]%
        {li2019adversarial}
\bibfield{author}{\bibinfo{person}{Juncheng Li}, \bibinfo{person}{Frank
  Schmidt}, {and} \bibinfo{person}{Zico Kolter}.}
  \bibinfo{year}{2019}\natexlab{}.
\newblock \showarticletitle{Adversarial camera stickers: A physical
  camera-based attack on deep learning systems}. In
  \bibinfo{booktitle}{\emph{International Conference on Machine Learning}}.
  PMLR, \bibinfo{pages}{3896--3904}.
\newblock


\bibitem[\protect\citeauthoryear{Li and Gal}{Li and Gal}{2017}]%
        {li2017dropout}
\bibfield{author}{\bibinfo{person}{Yingzhen Li} {and} \bibinfo{person}{Yarin
  Gal}.} \bibinfo{year}{2017}\natexlab{}.
\newblock \showarticletitle{Dropout inference in Bayesian neural networks with
  alpha-divergences}. In \bibinfo{booktitle}{\emph{International conference on
  machine learning}}. PMLR, \bibinfo{pages}{2052--2061}.
\newblock


\bibitem[\protect\citeauthoryear{Lin, Gan, and Han}{Lin et~al\mbox{.}}{2019}]%
        {lin2019defensive}
\bibfield{author}{\bibinfo{person}{Ji Lin}, \bibinfo{person}{Chuang Gan}, {and}
  \bibinfo{person}{Song Han}.} \bibinfo{year}{2019}\natexlab{}.
\newblock \showarticletitle{Defensive quantization: When efficiency meets
  robustness}.
\newblock \bibinfo{journal}{\emph{arXiv preprint arXiv:1904.08444}}
  (\bibinfo{year}{2019}).
\newblock


\bibitem[\protect\citeauthoryear{Madry, Makelov, Schmidt, Tsipras, and
  Vladu}{Madry et~al\mbox{.}}{2017}]%
        {madry2017towards}
\bibfield{author}{\bibinfo{person}{Aleksander Madry},
  \bibinfo{person}{Aleksandar Makelov}, \bibinfo{person}{Ludwig Schmidt},
  \bibinfo{person}{Dimitris Tsipras}, {and} \bibinfo{person}{Adrian Vladu}.}
  \bibinfo{year}{2017}\natexlab{}.
\newblock \showarticletitle{Towards deep learning models resistant to
  adversarial attacks}.
\newblock \bibinfo{journal}{\emph{arXiv preprint arXiv:1706.06083}}
  (\bibinfo{year}{2017}).
\newblock


\bibitem[\protect\citeauthoryear{Nicolae, Sinn, Tran, Buesser, Rawat, Wistuba,
  Zantedeschi, Baracaldo, Chen, Ludwig, et~al\mbox{.}}{Nicolae
  et~al\mbox{.}}{2018}]%
        {nicolae2018adversarial}
\bibfield{author}{\bibinfo{person}{Maria-Irina Nicolae},
  \bibinfo{person}{Mathieu Sinn}, \bibinfo{person}{Minh~Ngoc Tran},
  \bibinfo{person}{Beat Buesser}, \bibinfo{person}{Ambrish Rawat},
  \bibinfo{person}{Martin Wistuba}, \bibinfo{person}{Valentina Zantedeschi},
  \bibinfo{person}{Nathalie Baracaldo}, \bibinfo{person}{Bryant Chen},
  \bibinfo{person}{Heiko Ludwig}, {et~al\mbox{.}}}
  \bibinfo{year}{2018}\natexlab{}.
\newblock \showarticletitle{Adversarial Robustness Toolbox v1. 0.0}.
\newblock \bibinfo{journal}{\emph{arXiv preprint arXiv:1807.01069}}
  (\bibinfo{year}{2018}).
\newblock


\bibitem[\protect\citeauthoryear{Ovadia, Fertig, Ren, Nado, Sculley, Nowozin,
  Dillon, Lakshminarayanan, and Snoek}{Ovadia et~al\mbox{.}}{2019}]%
        {ovadia2019can}
\bibfield{author}{\bibinfo{person}{Yaniv Ovadia}, \bibinfo{person}{Emily
  Fertig}, \bibinfo{person}{Jie Ren}, \bibinfo{person}{Zachary Nado},
  \bibinfo{person}{David Sculley}, \bibinfo{person}{Sebastian Nowozin},
  \bibinfo{person}{Joshua~V Dillon}, \bibinfo{person}{Balaji Lakshminarayanan},
  {and} \bibinfo{person}{Jasper Snoek}.} \bibinfo{year}{2019}\natexlab{}.
\newblock \showarticletitle{Can you trust your model's uncertainty? Evaluating
  predictive uncertainty under dataset shift}.
\newblock \bibinfo{journal}{\emph{arXiv preprint arXiv:1906.02530}}
  (\bibinfo{year}{2019}).
\newblock


\bibitem[\protect\citeauthoryear{Papernot, McDaniel, Wu, Jha, and
  Swami}{Papernot et~al\mbox{.}}{2016}]%
        {papernot2016distillation}
\bibfield{author}{\bibinfo{person}{Nicolas Papernot}, \bibinfo{person}{Patrick
  McDaniel}, \bibinfo{person}{Xi Wu}, \bibinfo{person}{Somesh Jha}, {and}
  \bibinfo{person}{Ananthram Swami}.} \bibinfo{year}{2016}\natexlab{}.
\newblock \showarticletitle{Distillation as a defense to adversarial
  perturbations against deep neural networks}. In
  \bibinfo{booktitle}{\emph{2016 IEEE symposium on security and privacy (SP)}}.
  IEEE, \bibinfo{pages}{582--597}.
\newblock


\bibitem[\protect\citeauthoryear{Qendro, Chauhan, Ramos, and Mascolo}{Qendro
  et~al\mbox{.}}{2021}]%
        {qendro2021benefit}
\bibfield{author}{\bibinfo{person}{Lorena Qendro}, \bibinfo{person}{Jagmohan
  Chauhan}, \bibinfo{person}{Alberto Gil~CP Ramos}, {and}
  \bibinfo{person}{Cecilia Mascolo}.} \bibinfo{year}{2021}\natexlab{}.
\newblock \showarticletitle{The Benefit of the Doubt: Uncertainty Aware Sensing
  for Edge Computing Platforms}.
\newblock \bibinfo{journal}{\emph{arXiv preprint arXiv:2102.05956}}
  (\bibinfo{year}{2021}).
\newblock


\bibitem[\protect\citeauthoryear{Sharma and Chen}{Sharma and Chen}{2018}]%
        {sharma2018bypassing}
\bibfield{author}{\bibinfo{person}{Yash Sharma} {and} \bibinfo{person}{Pin-Yu
  Chen}.} \bibinfo{year}{2018}\natexlab{}.
\newblock \showarticletitle{Bypassing feature squeezing by increasing adversary
  strength}.
\newblock \bibinfo{journal}{\emph{arXiv preprint arXiv:1803.09868}}
  (\bibinfo{year}{2018}).
\newblock


\bibitem[\protect\citeauthoryear{Shumailov, Zhao, Mullins, and
  Anderson}{Shumailov et~al\mbox{.}}{2018}]%
        {shumailov2018taboo}
\bibfield{author}{\bibinfo{person}{Ilia Shumailov}, \bibinfo{person}{Yiren
  Zhao}, \bibinfo{person}{Robert Mullins}, {and} \bibinfo{person}{Ross
  Anderson}.} \bibinfo{year}{2018}\natexlab{}.
\newblock \showarticletitle{The taboo trap: Behavioural detection of
  adversarial samples}.
\newblock \bibinfo{journal}{\emph{arXiv preprint arXiv:1811.07375}}
  (\bibinfo{year}{2018}).
\newblock


\bibitem[\protect\citeauthoryear{Srivastava, Hinton, Krizhevsky, Sutskever, and
  Salakhutdinov}{Srivastava et~al\mbox{.}}{2014}]%
        {srivastava2014dropout}
\bibfield{author}{\bibinfo{person}{Nitish Srivastava},
  \bibinfo{person}{Geoffrey Hinton}, \bibinfo{person}{Alex Krizhevsky},
  \bibinfo{person}{Ilya Sutskever}, {and} \bibinfo{person}{Ruslan
  Salakhutdinov}.} \bibinfo{year}{2014}\natexlab{}.
\newblock \showarticletitle{Dropout: a simple way to prevent neural networks
  from overfitting}.
\newblock \bibinfo{journal}{\emph{The journal of machine learning research}}
  \bibinfo{volume}{15}, \bibinfo{number}{1} (\bibinfo{year}{2014}),
  \bibinfo{pages}{1929--1958}.
\newblock


\bibitem[\protect\citeauthoryear{Tram{\`e}r, Kurakin, Papernot, Goodfellow,
  Boneh, and McDaniel}{Tram{\`e}r et~al\mbox{.}}{2017}]%
        {tramer2017ensemble}
\bibfield{author}{\bibinfo{person}{Florian Tram{\`e}r}, \bibinfo{person}{Alexey
  Kurakin}, \bibinfo{person}{Nicolas Papernot}, \bibinfo{person}{Ian
  Goodfellow}, \bibinfo{person}{Dan Boneh}, {and} \bibinfo{person}{Patrick
  McDaniel}.} \bibinfo{year}{2017}\natexlab{}.
\newblock \showarticletitle{Ensemble adversarial training: Attacks and
  defenses}.
\newblock \bibinfo{journal}{\emph{arXiv preprint arXiv:1705.07204}}
  (\bibinfo{year}{2017}).
\newblock


\bibitem[\protect\citeauthoryear{Tsipras, Santurkar, Engstrom, Turner, and
  Madry}{Tsipras et~al\mbox{.}}{2018}]%
        {tsipras2018robustness}
\bibfield{author}{\bibinfo{person}{Dimitris Tsipras}, \bibinfo{person}{Shibani
  Santurkar}, \bibinfo{person}{Logan Engstrom}, \bibinfo{person}{Alexander
  Turner}, {and} \bibinfo{person}{Aleksander Madry}.}
  \bibinfo{year}{2018}\natexlab{}.
\newblock \showarticletitle{Robustness may be at odds with accuracy}.
\newblock \bibinfo{journal}{\emph{arXiv preprint arXiv:1805.12152}}
  (\bibinfo{year}{2018}).
\newblock


\bibitem[\protect\citeauthoryear{Xu, Liu, Li, and Tang}{Xu
  et~al\mbox{.}}{2020}]%
        {xu2020robust}
\bibfield{author}{\bibinfo{person}{Han Xu}, \bibinfo{person}{Xiaorui Liu},
  \bibinfo{person}{Yaxin Li}, {and} \bibinfo{person}{Jiliang Tang}.}
  \bibinfo{year}{2020}\natexlab{}.
\newblock \showarticletitle{To be Robust or to be Fair: Towards Fairness in
  Adversarial Training}.
\newblock \bibinfo{journal}{\emph{arXiv preprint arXiv:2010.06121}}
  (\bibinfo{year}{2020}).
\newblock


\bibitem[\protect\citeauthoryear{Xu, Evans, and Qi}{Xu et~al\mbox{.}}{2017}]%
        {xu2017feature}
\bibfield{author}{\bibinfo{person}{Weilin Xu}, \bibinfo{person}{David Evans},
  {and} \bibinfo{person}{Yanjun Qi}.} \bibinfo{year}{2017}\natexlab{}.
\newblock \showarticletitle{Feature squeezing: Detecting adversarial examples
  in deep neural networks}.
\newblock \bibinfo{journal}{\emph{arXiv preprint arXiv:1704.01155}}
  (\bibinfo{year}{2017}).
\newblock


\bibitem[\protect\citeauthoryear{Zhang, Goodfellow, Metaxas, and Odena}{Zhang
  et~al\mbox{.}}{2019}]%
        {zhang2019self}
\bibfield{author}{\bibinfo{person}{Han Zhang}, \bibinfo{person}{Ian
  Goodfellow}, \bibinfo{person}{Dimitris Metaxas}, {and}
  \bibinfo{person}{Augustus Odena}.} \bibinfo{year}{2019}\natexlab{}.
\newblock \showarticletitle{Self-attention generative adversarial networks}. In
  \bibinfo{booktitle}{\emph{International conference on machine learning}}.
  PMLR, \bibinfo{pages}{7354--7363}.
\newblock


\end{thebibliography}

\end{document}